# Is post-editing really faster than human translation?

Silvia Terribile

University of Manchester

## Abstract

Time efficiency is paramount for the localisation industry, which demands ever-faster turnaround times. However, translation speed is largely underresearched, and there is a lack of clarity about how language service providers (LSPs) can evaluate the performance of their post-editing (PE) and human translation (HT) services. This study constitutes the first large-scale investigation of translation and revision speed in HT and in the PE of neural machine translation, based on real-world data from an LSP. It uses an exploratory data analysis approach to investigate data for 90 million words translated by 879 linguists across 11 language pairs, over 2.5 years. The results of this research indicate that (a) PE is usually but not always faster than HT; (b) average speed values may be misleading; (c) translation speed is highly variable; and (d) edit distance cannot be used as a proxy for post-editing productivity, because it does not correlate strongly with speed.

## 1. Introduction

Despite the pressure for language service providers (LSPs) to work at an ever-faster pace, human translation (HT) speed is largely underresearched, and previous studies on post-editing (PE) speed have mostly focused on evaluating machine translation (MT) systems' performance, rather than the efficiency of PE itself (e.g. Läubli et al. 2019). This lack of research has severely affected the localisation industry's ability to adequately evaluate translation speed. Although numerous publications have reported significant time-savings through PE compared to HT (e.g. Kosmaczewska and Train 2019; Sánchez-Gijón, Moorkens and Way 2019), there are no widely accepted standards of how many words per hour (WPH) can typically be translated through these services. Previous research has usually analysed very limited data, and/or data collected under experimental conditions. There is large variability in the average translation speed values reported in different studies, and revision speed has largely been overlooked. Furthermore, some LSPs use edit distance – which measures technical effort, by calculating the number of characters that linguists edit to transform an MT output into a high-quality translation – to evaluate the productivity gains obtained through post-editing, and/or to determine linguists' remuneration (Albarino 2019; ELIA et al. 2022). However, previous studies have reported conflicting results about the correlation between technical and temporal PE effort (e.g. Cumbreño and Aranberri 2021; Moorkens et al. 2015).

This research constitutes the first large-scale investigation of speed in human translation and in the post-editing of neural machine translation (NMT), based on real-world data. It aims to enhance our understanding of a fundamental aspect of translation productivity, i.e. the speed at which human linguists are able to work, and to be a first step towards developing a more comprehensive approach to evaluating translation speed. It uses an exploratory data analysis approach to investigate the productivity reports of an LSP, TranslateMedia.[1] This examination focuses on data related to over 66 million words translated through HT, and almost 24 million words post-edited by 879 linguists, working across 11 language pairs, over a 2.5-year period. It considers speed at the translation and revision stages of HT and PE, aiming to shed some light on (a) whether and to what extent PE is faster than HT; (b) what ranges of speed are typical/atypical; (c) whether speed tends to be more variable in PE or HT; (d) how much variability there is in the average speed values achieved by different linguists; and (e) to what extent speed (temporal effort) correlates with edit distance (technical effort), at the translation stage of PE and at the revision stage of both PE and HT.

## 2. Related work

Professional translation blogs have widely discussed speed at the translation stage of HT. For example, according to PacTranz (2021), "[t]ranslator speeds and output vary enormously – anywhere from 200 to 500 words an hour", averaging at 300 WPH. Virino (2022) suggests that the same linguist may be able to translate from 150 to 600 WPH depending on the 'complexity' of the source and other contextual variables. To my knowledge, no professional publications have discussed PE speed, except for one forum discussion on ProZ.com (2018). This is surprising because, according to data from Memsource (2020), "post-editing has become the dominant translation method" in 2020 and an ever-growing number of LSPs, such as TranslateMedia, are using full post-editing – i.e. PE that aims to deliver a final translation of the highest possible quality – as an alternative to human translation.

---

[1] TranslateMedia was acquired by another LSP in 2022. This article refers to 'TranslateMedia', because the data analysed were created before the acquisition.

Scholars have widely discussed speed at the first round of both HT and PE, presenting very diverse average values. For instance, in an experiment on a novel translation, Toral, Wieling and Way (2018) reported an average of 503 WPH for HT, and 685 WPH for NMT post-editing. In the banking and finance domain, Läubli et al. (2019) recorded average values of 585 WPH in HT and 934 WPH in NMT post-editing in the German-to-French language pair, and 453 WPH in HT and 495 WPH in NMT post-editing in German to Italian. Numerous studies (e.g. Moorkens et al. 2015; Parra Escartín and Arcedillo 2015; Toral, Wieling and Way 2018) found large variance in speed among different linguists, due to their individual skills and ways of working. Macken, Prou and Tezcan (2020) compared the variance among the speed achieved by 20 linguists from the Directorate-General for Translation of the European Commission, reporting higher variability in PE (average: 410-910 WPH), than in HT (average: 450-720 WPH).

Revision speed has largely been disregarded: to my knowledge, no academic nor professional publications have discussed revision speed in HT, except for a few discussions on ProZ.com (2011; 2016a; 2016b). Moreover, since a major goal of PE is that of saving time compared to HT, much scholarship has assumed that no revision would be involved in the PE process. This is reflected also in the standards from the International Organization for Standardization (ISO): despite asserting that the output of full PE needs to be "comparable to a product obtained by human translation", revision is mentioned as a key step only for HT, and not for PE (British Standards Institute 2017, 2; cf. British Standards Institute 2015). Nevertheless, Temizöz (2017) has investigated revision speed in PE, measuring speed in post-editing carried out by a professional translator and revised by a subject-matter expert (non-professional linguist), and vice versa. These workflows significantly differ from those applied by most LSPs, such as TranslateMedia, where both translation and revision are carried out by professional linguists. The scope of Temizöz's research was limited to a 587-word technical text translated from English to Turkish by 20 participants (ibidem). She reported that, on average, linguists were able to post-edit 498 WPH and revise 1,752 WPH, whereas subject-matter experts post-edited 546 WPH and revised 1,116 WPH[2] (ibidem).

To measure PE effort and evaluate any gains compared to HT, much scholarship has considered three dimensions, defined in Krings' (2001) seminal work as temporal, technical, and cognitive effort. Temporal effort represents the time needed for post-editing (ibidem). LSPs typically consider time in relation to throughput, thus focusing on speed, expressed as the words per hour ratio. Technical effort involves all the operations needed to practically carry out PE, namely "deletion, insertion, reordering, or a combination of these operations" (ibid., 179). Among other methods, technical effort is frequently measured through edit distance. Finally, cognitive effort is defined as "the type and extent of those cognitive processes" activated during post-editing (ibid., 179). It is the most challenging to measure, because only indirect measurements are possible (ibidem).

---

[2] Values here converted from words per minute (WPM) to WPH.

Despite the complexity of measuring PE effort, some LSPs use edit distance as a key proxy for post-editing productivity, and/or to determine linguists' remuneration (Albarino 2019; ELIA et al. 2022). However, cognitive effort is at times only weakly correlated to temporal and technical effort (Krings 2001; Moorkens et al. 2015). If we focus on the types of effort that are the most feasible for LSPs to consider, i.e. temporal and technical effort, to my knowledge, only four small-scale studies have investigated their interrelation. In the post-editing of statistical machine translation (SMT), O'Brien (2011, 19) found that speed and edit distance "correlate well", and Moorkens et al. (2015) identified strong correlations between these variables. Macken, Prou and Tezcan (2020) reported weak to moderate associations between speed and edit distance in the PE of SMT, and moderate in the PE of NMT, and Cumbreño and Aranberri (2021) found only weak correlations in NMT PE. No previous research has investigated this correlation at the revision stage of PE or HT.

### 3. Data overview and selection

TranslateMedia offers a wide range of language services, including human translation and full post-editing of adaptive NMT – here referred to simply as post-editing – in which post-editors edit NMT outputs to achieve translations of publishable quality, and the NMT engine adapts and learns from their corrections (Kosmaczewska and Train 2019). HT and PE are carried out within the same computer-assisted translation (CAT) environment, a customised version of memoQ. When available, translation memory (TM) matches are provided for all segments in both HT and PE jobs.

Quality expectations are the same for HT and PE services, as both aim to deliver final translations of the highest possible quality, which will be published by clients. To this aim, all HT and PE work consists of a first round of translation, followed by a revision stage carried out by a second human linguist, who proofreads and makes any corrections to the target text. As such, this article refers to a 'translation stage' and a 'revision stage', in the context of both human translation and post-editing tasks. All linguists are professional freelance translators and translate into their native language. They have different backgrounds and levels of experience, and work remotely across the globe. Most post-editors (92% in the analysed dataset) work on HT too; since the volume of texts translated through HT is much higher than that of post-edited texts, only 22% of linguists working on HT provide PE as well. The LSP's project and account managers liaise with their clients to determine whether HT or PE may be the most suitable service for each project on a case-by-case basis, considering clients' needs and MT quality in different domains and language combinations, among other factors.

TranslateMedia gathers quantitative productivity data, which help them manage and evaluate their performance. These data include some productivity measures (e.g. translation speed and edit distance) as well as data related to other factors (e.g. language pair, text genre, etc.) that can affect productivity. The LSP provided me with PE and HT data presented from different perspectives – i.e. data by HT/PE job, by language pair, by translator/post-editor, and by reviser – which have been useful to investigate various aspects of translation speed. These data were saved into an Excel workbook format (.xlsx), and processed through the statistical analysis software IBM SPSS (Statistical Package for the Social Sciences), version 25 (IBM 2021).

This investigation started by considering data for all PE and HT tasks in all available language pairs completed from January 2019 to mid-June 2021, when this research began. 2019 was chosen as the starting year because the LSP adopted NMT in 2018, but only a limited number of texts were post-edited using NMT that year. A preliminary analysis demonstrated that the amount of work carried out in some language pairs was not sufficient to achieve reliable results. Therefore, further data selection was undertaken through an empirically driven process based on visualisations of translation speed values – not discussed here due to spatial constraints – resulting in the selection of 11 language pairs with the highest number of translated words. The selected language pairs are English (EN) to Danish (DA) / Dutch (NL) / Finnish (FI) / French (FR) / German (DE) / Italian (IT) / Polish (PL) / Portuguese (PT) / Spanish (ES) / Swedish (SV)[3], and French to English. The majority of texts translated in all these language pairs belonged to the genres of marketing (average: 63% in HT, 86% in PE) or technical marketing, i.e. marketing texts with a high percentage of technical terminology (average: 22% in HT, 12% in PE). Table 1 summarises the quantities of analysed data, and Table 2 displays the number of linguists.

---

[3] Language labels from ISO 639-1 (Library of Congress 2017).

| | Quantities of data | | | |
|---|---|---|---|---|
| | Human translation | | Post-editing | |
| **Language pair** | **Jobs** | **Words** | **Jobs** | **Words** |
| EN>DA | 1,007 | 2,194,805 | 639 | 1,043,931 |
| EN>DE | 6,813 | 14,544,521 | 3,172 | 4,765,551 |
| EN>ES | 4,891 | 9,869,196 | 1,349 | 3,250,709 |
| EN>FI | 860 | 1,242,069 | 789 | 1,274,238 |
| EN>FR | 6,021 | 11,668,452 | 2,248 | 3,936,132 |
| EN>IT | 6,151 | 10,884,517 | 1,266 | 2,212,146 |
| EN>NL | 2,724 | 5,168,568 | 1,167 | 2,916,506 |
| EN>PL | 2,336 | 3,710,768 | 1,098 | 1,818,600 |
| EN>PT | 2,075 | 4,218,121 | 237 | 587,994 |
| EN>SV | 890 | 2,005,279 | 588 | 1,008,192 |
| FR>EN | 492 | 975,566 | 88 | 922,288 |
| **Total** | 34,260 | 66,481,862 | 12,641 | 23,736,287 |

**Table 1.** Quantities of data analysed in this research

| | Number of linguists (translation + revision) | | | |
|---|---|---|---|---|
| **Language pair** | **HT + PE** | **Only HT** | **Only PE** | **Total** |
| EN>DA | 8 | 15 | 2 | 25 |
| EN>DE | 19 | 131 | 0 | 150 |
| EN>ES | 18 | 98 | 3 | 119 |
| EN>FI | 14 | 14 | 3 | 31 |
| EN>FR | 22 | 145 | 0 | 167 |
| EN>IT | 20 | 79 | 1 | 100 |
| EN>NL | 20 | 60 | 0 | 80 |
| EN>PL | 23 | 28 | 0 | 51 |
| EN>PT | 22 | 38 | 1 | 61 |
| EN>SV | 14 | 16 | 0 | 30 |
| FR>EN | 17 | 41 | 7 | 65 |
| **Total** | 197 | 665 | 17 | 879 |

**Table 2.** Number of linguists

Among all variables included in TranslateMedia's productivity reports, this research has focused especially on translation speed. WPH values are automatically recorded by the LSP's CAT tool, and they represent the source wordcount, divided by the time a linguist spends in the CAT environment. When there is no activity within the CAT for over two minutes, all minutes of inactivity are deleted, ensuring that breaks are not included. However, this also means that time spent researching and reading any briefing documents for over two minutes is not considered. Besides, some of these values may be incorrect, mostly due to technical issues or to linguists occasionally working offline. Overall, whilst these WPH values are not as precise as if they were collected under experimental conditions, they were an invaluable resource, as they captured speed in a very large number of translation tasks, in a wide range of real-world contexts.

Edit distance values have also been essential for this research. They are measured automatically in the LSP's CAT through the Levenshtein algorithm, which "calculates the minimum number of character edits that are necessary to transform one string into another string" (Kosmaczewska and Train 2019, 170). This algorithm is rather complex, and it is beyond the scope of this article to discuss it. For clarity, I would argue that this simplified equation provides a basic way to calculate the Levenshtein distance: Edit distance = minimum number of edited characters / number of reference characters. This metric is expressed as a percentage: the lower the percentage, the fewer the edits. This research has examined three types of edit distance: post-editing distance (PED), PED including TM matches and repetitions – which I will refer to as 'PED-TMR' – and revision distance. PED considers corrections to the MT output only, and it is frequently used to evaluate MT systems' performance. PED-TMR accounts for the overall technical PE effort. Indeed, post-editors edit a combination of MT output and TM matches; in case of repetitions, linguists edit the segment once, and their edits are auto-propagated. Finally, revision distance is measured for both PE and HT tasks, and it calculates the distance between the translated and revised texts.

## 4. Exploratory data analysis approach

This research has applied an exploratory data analysis approach, which involved using descriptive statistics to explore TranslateMedia's productivity data (Cleff 2014). The distribution of speed data was tested to understand whether parametric statistics – which assume that the data under investigation follow a normal distribution – would be appropriate for this study (Mellinger and Hanson 2016). The Kolmogorov-Smirnov (K-S) test of normality was used, as it is the most suitable test for datasets with over 50 data points (Pennsylvania State University 2022). The K-S test produces (1) a distance (D) value, which ranges from 0 to 1 and represents the maximum distance between the analysed data and the normal distribution; and (2) a probability value (p-value): if this is lower than .05, there is sufficient evidence to reject the null hypothesis according to which the analysed data are normally distributed (Mellinger and Hanson 2016). Table 3 presents the results of the Kolmogorov-Smirnov test of WPH by HT/PE job at the translation stage, and Table 4 displays the same information from the revision stage.

| Kolmogorov-Smirnov normality test of WPH by HT/PE job, at the translation stage | | | | | | | |
|---|---|---|---|---|---|---|---|
| HT | | | | PE | | | |
| W/ outliers | | W/o outliers | | W/ outliers | | W/o outliers | |
| **Language pair** | D | p | D | p | D | p | D | p |
| EN>DA | 1 | < .001 | 1 | < .001 | 1 | < .001 | 1 | < .001 |
| EN>DE | 1 | < .001 | 1 | < .001 | .99966 | < .001 | .99964 | < .001 |
| EN>ES | .99979 | < .001 | .99979 | < .001 | 1 | < .001 | 1 | < .001 |
| EN>FI | 1 | < .001 | 1 | < .001 | .99746 | < .001 | .99734 | < .001 |
| EN>FR | 1 | < .001 | 1 | < .001 | 1 | < .001 | 1 | < .001 |
| EN>IT | .99980 | < .001 | .99980 | < .001 | 1 | < .001 | 1 | < .001 |
| EN>NL | 1 | < .001 | 1 | < .001 | 1 | < .001 | 1 | < .001 |
| EN>PL | .99997 | < .001 | .99997 | < .001 | 1 | < .001 | 1 | < .001 |
| EN>PT | .99948 | < .001 | .99947 | < .001 | 1 | < .001 | 1 | < .001 |
| EN>SV | 1 | < .001 | 1 | < .001 | 1 | < .001 | 1 | < .001 |
| FR>EN | 1 | < .001 | 1 | < .001 | 1 | < .001 | 1 | < .001 |
| All pairs | .99988 | < .001 | .99987 | < .001 | .99974 | < .001 | .99973 | < .001 |

**Table 3.** Kolmogorov-Smirnov normality test of WPH by HT/PE job, at the translation stage

| Kolmogorov-Smirnov normality test of WPH by HT/PE job, at the revision stage | | | | | | | |
|---|---|---|---|---|---|---|---|
| HT | | | | PE | | | |
| W/ outliers | | W/o outliers | | W/ outliers | | W/o outliers | |
| **Language pair** | D | p | D | p | D | p | D | p |
| EN>DA | 1 | < .001 | 1 | < .001 | 1 | < .001 | 1 | < .001 |
| EN>DE | 1 | < .001 | 1 | < .001 | 1 | < .001 | 1 | < .001 |
| EN>ES | .99997 | < .001 | .99997 | < .001 | 1 | < .001 | 1 | < .001 |
| EN>FI | 1 | < .001 | 1 | < .001 | 1 | < .001 | 1 | < .001 |
| EN>FR | 1 | < .001 | 1 | < .001 | 1 | < .001 | 1 | < .001 |
| EN>IT | 1 | < .001 | 1 | < .001 | 1 | < .001 | 1 | < .001 |
| EN>NL | 1 | < .001 | 1 | < .001 | 1 | < .001 | 1 | < .001 |
| EN>PL | 1 | < .001 | 1 | < .001 | 1 | < .001 | 1 | < .001 |
| EN>PT | 1 | < .001 | 1 | < .001 | 1 | < .001 | 1 | < .001 |
| EN>SV | 1 | < .001 | 1 | < .001 | 1 | < .001 | 1 | < .001 |
| FR>EN | 1 | < .001 | 1 | < .001 | 1 | < .001 | 1 | < .001 |
| All pairs | .99997 | < .001 | .99997 | < .001 | 1 | < .001 | 1 | < .001 |

**Table 4.** Kolmogorov-Smirnov normality test of WPH by HT/PE job, at the revision stage

As can be seen above, the null hypothesis was rejected in all cases. Since normality tests may sometimes produce inaccurate results especially when dealing with large datasets (Mellinger and Hanson 2016), the distribution of speed values was cross-checked through data visualisations – namely normal quantile-quantile plots, not presented here due to spatial constraints – which confirmed that the observed data are not normally distributed. As such, nonparametric statistical tests were used in this research.

To gain a general understanding of whether PE is usually faster than HT, this project started by considering average speed by language pair. However, average values can only provide a very partial picture of speed, especially because they are not necessarily representative of how individual values are spread out (Cleff 2014). Firstly, outliers, i.e. extreme cases, may significantly affect averages – as well as standard deviations and correlations, also examined in this study. Since some WPH values may be incorrect, it was fundamental to identify outliers, and check whether they constitute valid cases.

Previous research dealing with translation speed (e.g. Parra Escartín and Arcedillo 2015) has typically investigated small datasets and identified outliers by establishing thresholds of possible/impossible values. However, such an approach may not be effective in revealing trends and patterns in large datasets. Therefore, here outliers were identified through the mathematical formula applied in SPSS, which uses quartiles to consider the distribution of individual values in the dataset. The first quartile (Q1) is the value under which 25% of cases are found, the third quartile (Q3) is the value under which 75% of cases are situated, and the difference between Q3 and Q1 constitutes the interquartile range (IQR) (Cleff 2014). Outliers are identified through the following equation: Outliers =/< Q1 – (1.5 × IQR) or =/> Q3 + (1.5 × IQR) (Forsyth 2018). Quartiles have also been helpful to identify the ranges of the most common translation speed values. Indeed, the range between the first and third quartile comprises the middle 50% of HT/PE jobs, which may be considered as the most typical values.

Additionally, this project has investigated speed variability**,** aiming to shed some light on whether the speed obtained in different tasks is more variable in PE or HT, and how much variability there is in the average speed achieved by different linguists, due to their individual skills and ways of working. Previous research (e.g. Macken, Prou and Tezcan 2020) has typically considered the distance between the minimum and maximum average speed values obtained in different contexts as a proxy for speed variability. However, these ranges are not necessarily representative of how values are distributed. Therefore, here speed variability was explored through the interquartile range and standard deviation (SD). The IQR provides an indication of the extent to which values in the middle 50% of the data are spread out: the higher the IQR, the higher the dispersion in this range. It has the advantage of being resistant to outliers (Mellinger and Hanson 2016), but the disadvantage of considering only the 50% of data around the mean. To consider the distribution of all speed values, this study has also looked at the standard deviation, which measures the average distance between individual values and the mean: the higher the standard deviation, the higher the variability (Cleff 2014). Since standard deviation may be highly affected by outliers, the SD values with and without outliers were compared. Due to spatial constraints, in this article IQR values will be discussed in greater detail than SD ones.

Finally, the correlation between translation speed and edit distance was investigated, to understand to what extent temporal and technical effort are interrelated, and whether edit distance can be used as a proxy for productivity. A correlation can be positive or negative: when positive, if one variable grows, the other grows too, and vice versa; when negative, if one variable grows, the other decreases (Forsyth 2018). The strength of all correlations has been measured through the Kendall's tau correlation coefficient, as much research has reported that it is more powerful than other nonparametric correlation coefficients, such as Spearman's rho (Mellinger and Hanson 2016). In particular, this project has used Kendall's tau-b, which corrects for ties in the data. Kendall's tau-b may range from -1 to +1: a value of -1 indicates a perfect negative correlation; 0 signals that there is no linear relationship between the variables; and 1 represents a perfect positive correlation (ibidem). In this research, a coefficient in the range of .06 to .25 has been interpreted as a weak correlation, a value between .26 and .48 as a moderate correlation, a value between .49 to .70 as a strong correlation, and a value equal to or larger than .71 as a very strong correlation (Wicklin 2023). The p-value indicates how likely it is that the correlation coefficient is incorrect – for example, because the analysed values seem to follow a pattern by coincidence (Forsyth 2018). A correlation is statistically significant if the p-value is equal to or under .05 – indicating an error probability of 5% – and it is highly significant if it is equal to or lower than .01 (error probability: 1%) (ibidem).

When examining correlations excluding outliers, only translation speed outliers were disregarded. Indeed, in the analysed data, all edit distance outliers appear to be valid. In particular, (a) none of these values look 'impossible', as none are above 100%; (b) they are less subject to technical or measurement issues. For example, if a linguist works offline, this will affect the translation speed value, but not edit distance; and (c) any doubts about their validity can be solved by manually calculating edit distance.

## 5. Results

This section discusses the results of this research, which aimed to understand which translation and revision speed values are typical/atypical in PE and HT (Subsection 5.1), how variable speed is (Subsection 5.2), and to what extent temporal and technical effort are correlated (Subsection 5.3).

*5.1. (A)typical translation speed values*

Figure 1 displays average translation and revision WPH in HT and PE (average calculated by HT/PE job). In these and other figures in this article, the darkest colour indicates the addition of PE to the HT values, the percentages above bars indicate the speed increase through PE, and the value that accounts for all language pairs is highlighted in bold.

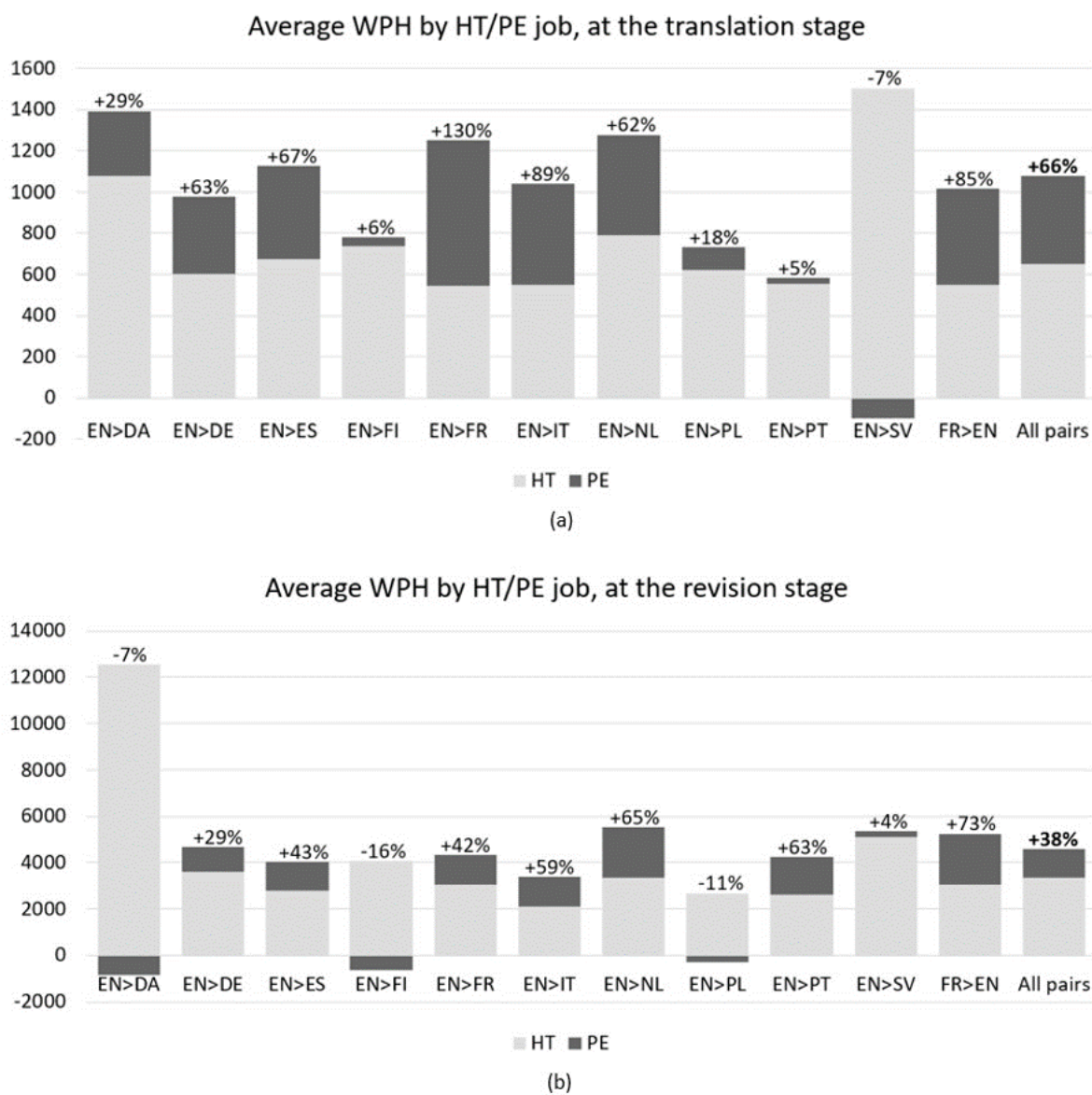

**Figure 1.** Stacked bar chart of average WPH by HT/PE job, at the (a) translation and (b) revision stages

These values are higher than those reported in previous publications. However, as previously mentioned, they do not include time spent reading briefing documents, researching, and taking breaks. At the translation stage, on average PE has greatly enhanced time efficiency compared to HT – i.e. it has considerably increased the amount of work completed in the same period of time – with overall time-savings of 66%. However, the impact of post-editing on speed has varied dramatically among different language pairs, from an average increase of +130% in English to French, to a decrease of -7% in English to Swedish. This shows that PE may not always increase speed in all language combinations, even when a significant amount of post-editing work has been completed in a language pair, since over a million words were translated from English to Swedish through post-editing.

Average revision values were also higher in PE compared to HT in 8 out of the 11 language pairs, with an overall speed increase of +38%. It is surprising to see such a striking difference between HT and PE at the revision stage, because (a) the same quality standards apply to the translation and revision of both HT and PE, and (b) the revision process is not supposed to present significant differences whether carried out as part of a PE or HT workflow; accordingly, linguists are paid the same for these services by TranslateMedia. I hypothesise that some of the possible reasons behind this result may be that (1) although linguists are requested to apply the same quality standards to HT and PE, some of them might perceive the post-editing practice as inferior to the human translation practice, which could lead them to be less accurate and complete PE revision more quickly; and/or (2) the quality of PE outputs might be frequently higher than that of HT outputs, enabling linguists to work faster on PE revision. If the latter were the case, it could be due to various reasons. For example, since the source texts processed through PE at TranslateMedia are typically "well-structured [and] often repetitive" texts (Kosmaczewska and Train 2019, 169), it would be possible that they tend to be more straightforward to translate than those usually processed through HT, facilitating the work of linguists on both rounds of PE. Nevertheless, testing these hypotheses would require an experimental design, which is beyond the scope of this exploratory data analysis.

In any case, as previously mentioned, average values only provide a partial picture of speed, and they may be affected by outliers. Figure 2 compares average translation speed, including and excluding outliers.

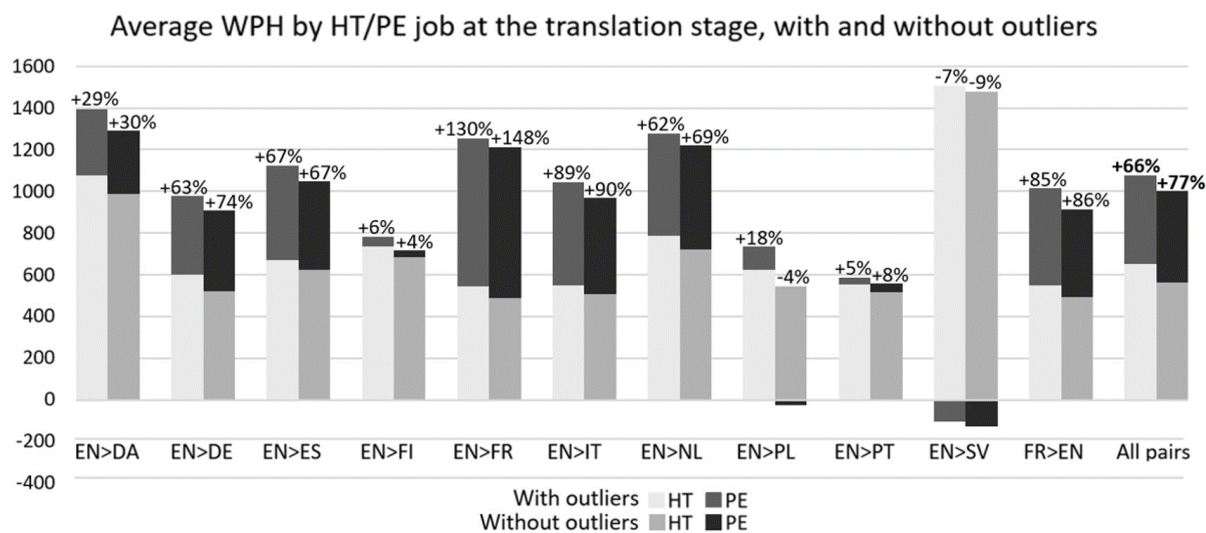

**Figure 2.** Stacked bar chart of average translation WPH by HT/PE job, with and without outliers

When excluding outliers, the overall average WPH values decrease by 13% in HT and by 7% in PE. However, results vary considerably among language pairs, from a difference of just -2% in English-to-Swedish HT, to -29% in English-to-Polish PE. In the latter case, the mean values that include outliers would suggest that, on average, linguists have saved 18% of their time through post-editing. Conversely, the mean values without outliers indicate that, on average, linguists took 4% longer when post-editing. Before disregarding any outliers, it is fundamental to investigate whether these values are invalid, or there could be some reasons explaining their presence. To gain a general idea of whether these values are realistic, I have investigated the highest, middle, and lowest values within all outliers for each language pair, both in PE and HT. In most cases, I could not exclude that these values were impossible *a priori*.

For example, in English-to-Polish post-editing, the analysed outliers were: 9,721 WPH, 2,378 WPH, and 1,709 WPH. At the bare minimum, a post-editing task involves reading the source text and the MT output, and average reading speed ranges from 200 to 330 WPM (Rayner, Slatter and Bélanger 2010). Although no previous research has investigated reading speed in post-editing or other professional translation contexts, I would argue that linguists are likely to read at a slower pace, because they need to be very careful to spot mistakes and inconsistencies. Nevertheless, these ranges may help us identify unrealistically high speed values. If a post-editor were able to read at the highest speed in this range, they could theoretically read up to 19,800 WPH; if we halve this value as they would read both the source and target texts (although their lengths may differ), they could read up to 9,900 words in an hour. In the highest outlier for English-to-Polish PE – i.e. 9,721 WPH – the post-editor has not only read the source and target texts, but they have also made some edits (PED-TMR:

19.52%). Thus, we can be confident that this WPH value is invalid, and it was likely caused by measurement or technical issues. However, the other two outliers are much lower than 9,900 WPH, so they could potentially be valid. I searched for any possible causes that would explain their validity, by looking into any available resources, such as the TM, the number of edits made during translation and revision, any localisation briefs, etc. However, the results of this analysis were inconclusive.

These outliers are likely to include a combination of valid and invalid cases. Therefore, when examining values that may be affected by outliers, I decided to (1) consider speed values both including and excluding outliers, and (2) if trends emerge from the data without outliers, calculate what percentage of all cases the data without outliers comprise, to understand whether the results obtained by disregarding extreme values relate to a considerable share of all data. In the case of English-to-Polish PE speed, even if all outliers were valid, PE would have been very time-efficient in these cases (average speed in the English-to-Polish PE outliers: 2,522 WPH), which are only 11% of jobs. Conversely, on average PE has been counterproductive in the remaining 89% of tasks. Indeed, in these cases the average PE speed is 520 WPH, which is lower than the average speed achieved through HT, both including outliers (622 WPH), and excluding them (544 WPH).

At the revision stage, the number of outliers is even higher than in translation, as shown in Figure 3.

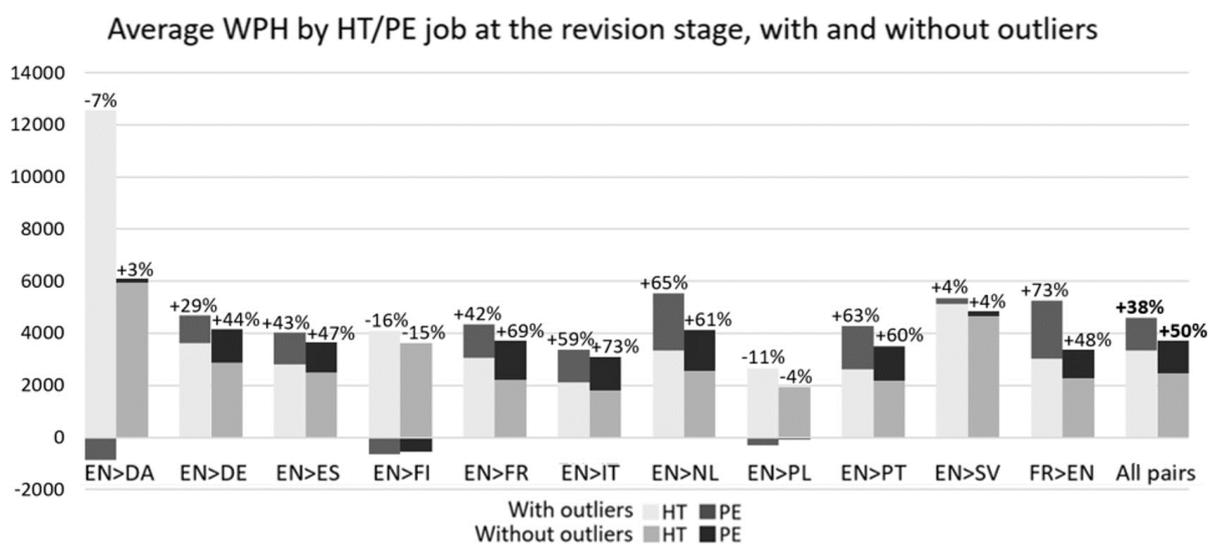

**Figure 3.** Stacked bar chart of average revision WPH by HT/PE job, with and without outliers

When disregarding outliers, we sometimes obtain a large difference in the revision speed increase achieved through PE. However, in all pairs other than English to Danish, this difference would never reverse the speed increase value from positive to negative, or vice versa. As such, it is not particularly relevant to calculate the percentage of time-efficient jobs in PE and/or HT. By contrast, the speed increase/decrease has changed from -7% to +3% in English to Danish. Here, the average revision speed values including outliers are extremely high, both in HT and PE, but they have roughly halved after excluding outliers. Since, at the bare minimum, a revision task involves reading the target text, based on the previously mentioned reading speed ranges, theoretically it would be possible to revise up to 19,800 WPH. Due to the large amount of extremely high speed values in English-to-Danish revision, the thresholds for identifying outliers were very high in this language pair, i.e. outliers =/> 20,081 WPH in HT, and 19,526 WPH in PE. Thus, here all revision outliers were roughly above the theoretical limit of 19,800 words that could potentially be revised in an hour and, as such, they are very likely to be invalid. Consequently, we can conclude that the average revision speed values including outliers are unreliable in this language pair, and those without outliers are considerably more realistic.

Finally, considering quartiles has enabled me to identify the ranges of the most common speed values. In Table 5, these figures are rounded to the nearest multiple of ten for the sake of memorability.

| Language pair | Ranges of the most typical WPH values by HT/PE job | | | |
| | Translation stage | | Revision stage | |
| | HT | PE | HT | PE |
|---|---|---|---|---|
| EN>DA | 530-1,420 | 810-1,800 | 3,440-10,090 | 4,400-10,450 |
| EN>DE | 310-750 | 480-1,310 | 1,420-4,430 | 2,080-6,110 |
| EN>ES | 350-890 | 650-1,470 | 1,270-3,620 | 2,160-5,220 |
| EN>FI | 470-910 | 490-980 | 1,790-5,440 | 1,870-4,390 |
| EN>FR | 320-680 | 800-1,610 | 1,140-3,450 | 2,130-5,350 |
| EN>IT | 290-720 | 500-1,420 | 860-2,730 | 1,730-4,380 |
| EN>NL | 440-1,010 | 680-1,740 | 1,390-3,920 | 2,330-6,380 |
| EN>PL | 270-870 | 290-860 | 840-3,240 | 1,040-2,820 |
| EN>PT | 330-710 | 390-750 | 1,300-3,180 | 1,940-5,260 |
| EN>SV | 780-2,080 | 550-2,070 | 2,320-6,940 | 2,820-6,860 |
| FR>EN | 270-710 | 740-1,150 | 890-3,860 | 1,540-6,060 |
| All pairs | **330-820** | **530-1,440** | **1,200-3,870** | **1,990-5,540** |

**Table 5.** Ranges of the most typical WPH values by HT/PE job (Q1-Q3) (overall values in bold)

### 5.2. Is speed more variable in post-editing or human translation?

To compare the variability of speed values obtained at the translation and revision stages of HT and PE, Figure 4 shows the interquartile range among WPH by HT/PE job.[4] For ease of reference and readability, the values related to all language pairs are also presented numerically in Table 6, together with the standard deviation values related to the same data, for the sake of increased robustness.

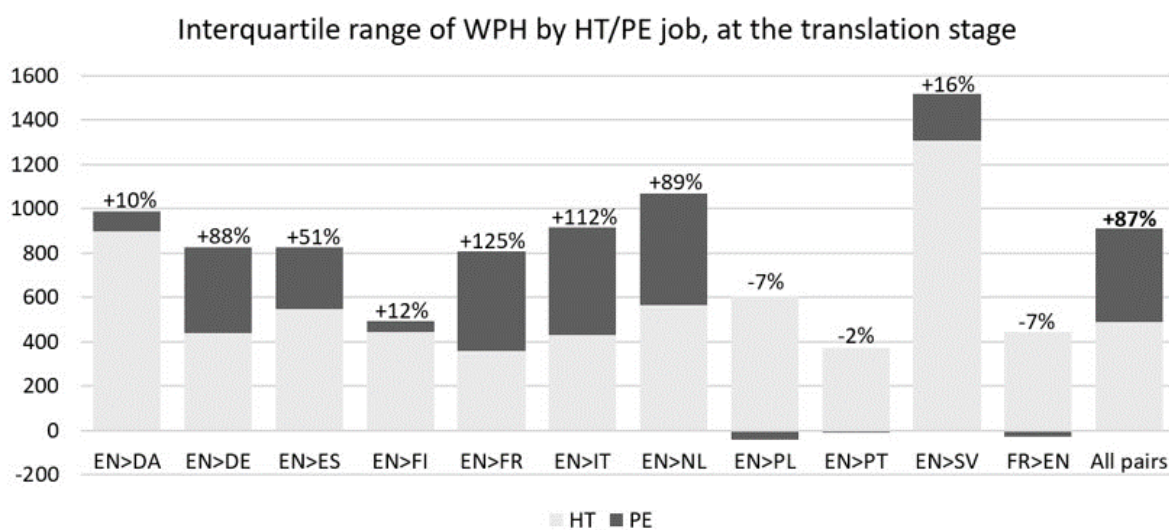

(a)

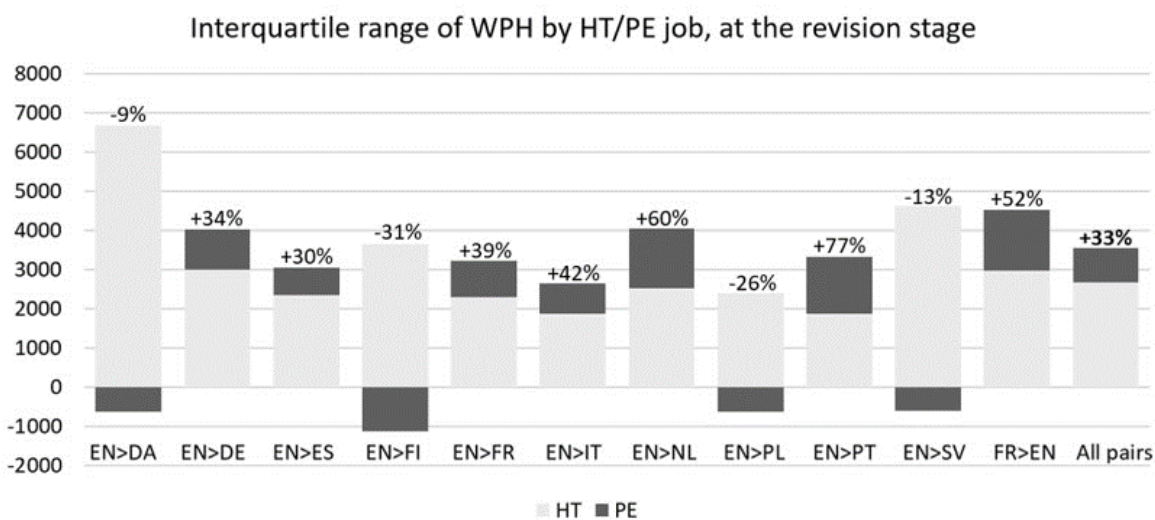

(b)

**Figure 4.** Stacked bar chart of interquartile range of WPH by HT/PE job, at the (a) translation and (b) revision stages

---

[4] Only values including outliers are presented, as the IQR is resistant to outliers (Mellinger and Hanson 2016).

| | Overall dispersion of WPH by HT/PE job | | | | | |
|---|---|---|---|---|---|---|
| | Translation stage | | | Revision stage | | |
| | HT | PE | PE-HT (%) | HT | PE | PE-HT (%) |
| IQR | 488 WPH | 911 WPH | +87% | 2,670 WPH | 3,557 WPH | +33% |
| SD w/ outliers | 519 WPH | 747 WPH | +44% | 8,974 WPH | 5,306 WPH | -41% |
| SD w/o outliers | 326 WPH | 608 WPH | +87% | 1,722 WPH | 2,319 WPH | +35% |

**Table 6.** Overall dispersion of WPH by HT/PE job

As seen above, there was great variability in the number of words processed in one hour, with sometimes high differences across language pairs. At the translation stage, the interquartile range is overall 87% higher in PE than in HT. In revision, the IQR values are extremely high, and they are on average 33% higher in PE compared to HT. The standard deviation values without outliers, which account for the vast majority of all data – i.e. 95% of translation data and 93% of revision ones in HT, and 97% of post-editing data and 95% of revision ones in PE – provide a qualitatively similar picture: SD is much higher in revision compared to translation, it is overall 87% higher in PE than HT at the translation stage, and 35% higher in PE compared to HT in revision.

To gain a fuller picture of variability in temporal effort, in addition to evaluating dispersion in the number of words translated in an hour, I would argue that it would be beneficial for LSPs to consider variability also in the amount of time spent to process a text of a given number of words. Indeed, this would be highly relevant especially for LSPs outsourcing work to freelance translators – which constitute the vast majority of LSPs (Moorkens 2020) – as they commission texts to translate rather than hours of translation work to their linguists, so it would be useful for them to understand how translation time varies across the translation and revision stages of human translation and post-editing, given a text of a certain wordcount.

Therefore, in addition to evaluating variability in the number of words processed in a fixed amount of time (here, an hour), I decided to consider dispersion also in the number of hours spent to process a fixed number of words – which I set to 1,000 words. In particular, I converted all speed values from WPH to hours per 1,000 words (H/KW) – calculated as (1/WPH) × 1000 – and I measured dispersion in the H/KW values as well.[5] Figure 5 displays

---

[5] It may be argued that the interquartile range and standard deviation of the H/KW values are likely to go exactly in the opposite direction of the IQR and SD of WPH. However, please note that converting speed values from WPH to H/KW involves carrying out a non-linear transformation. As such, it was necessary to verify the extent to which the IQR and SD of the H/KW values would present an opposite picture compared to their corresponding measures of dispersion related to the WPH values. Indeed, the IQR and SD of the H/KW values displayed in Figure 5 and Table 7 are considerably different from the values obtained by converting the IQR and SD values from WPH to H/KW (not presented here for the sake of conciseness).

the interquartile range of H/KW by HT/PE job, and Table 7 presents the overall IQR and SD
values.

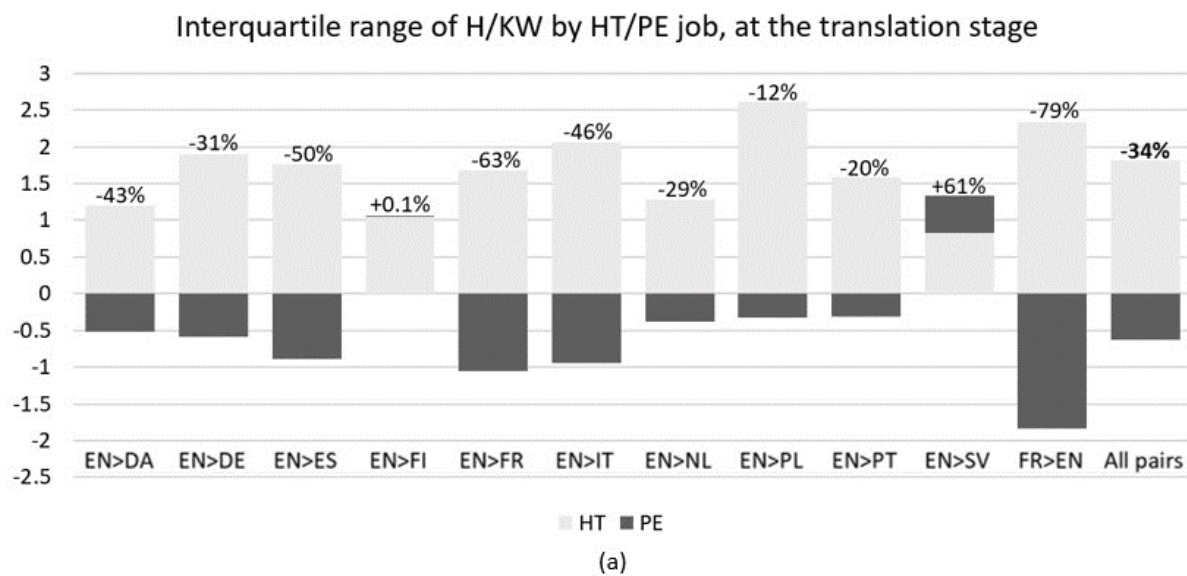

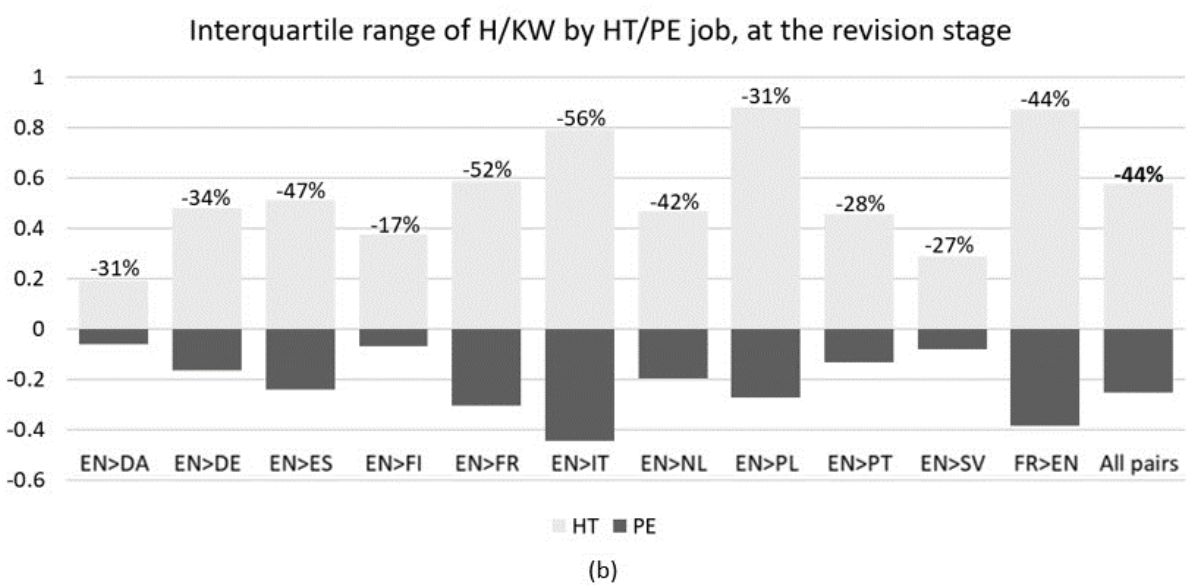

**Figure 5.** Stacked bar chart of interquartile range of H/KW by HT/PE job, at the (a) translation and (b)
revision stages

| | Overall dispersion of H/KW by HT/PE job | | | | | |
|---|---|---|---|---|---|---|
| | Translation stage | | | Revision stage | | |
| | HT | PE | PE-HT (%) | HT | PE | PE-HT (%) |
| IQR | 1.82 H/KW | 1.19 H/KW | -34% | 0.57 H/KW | 0.32 H/KW | -44% |
| SD w/ outliers | 8.42 H/KW | 11.05 H/KW | +31% | 1.99 H/KW | 1.23 H/KW | -38% |
| SD w/o outliers | 1.17 H/KW | 0.76 H/KW | -35% | 0.37 H/KW | 0.21 H/KW | -43% |

**Table 7.** Overall dispersion of H/KW by HT/PE job

These values show the other side of the coin: H/KW dispersion is overall considerably lower in revision than in translation, and much lower in PE compared to HT. On average, IQR values are 34% lower in PE than in HT at the translation stage, and 44% lower in PE compared to HT in revision. At the translation stage, the standard deviation of H/KW is highly affected by a great number of outliers, resulting in an overall SD that is 31% higher in PE than in HT. However, standard deviations without outliers – which account for 59% of PE data and 68% of HT ones at the translation stage – are qualitatively similar to the IQR values, as SD is 35% lower in PE than in HT. At the revision stage, the IQR values are qualitatively confirmed by the SD values both including and excluding outliers.

To shed some light on the extent to which the skills and ways of working of individual translators impact their speed, Figure 6 presents the interquartile range of average WPH and average H/KW achieved by different linguists working at the translation stage of PE and HT. Please see Table 8 for a summary of the overall IQR and SD values.

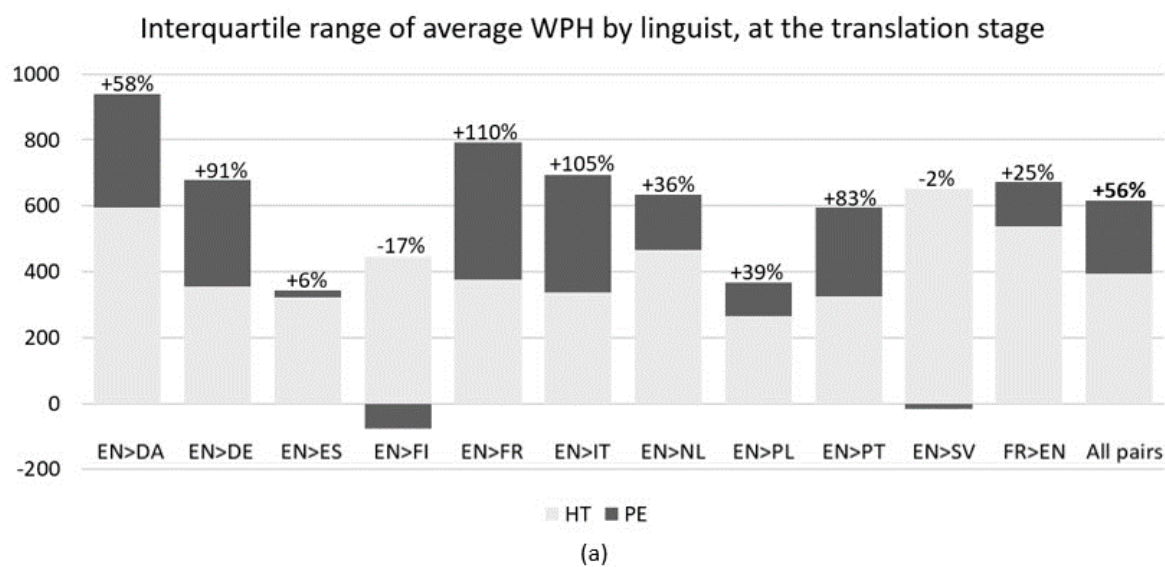

Interquartile range of average WPH by linguist, at the translation stage

(a)

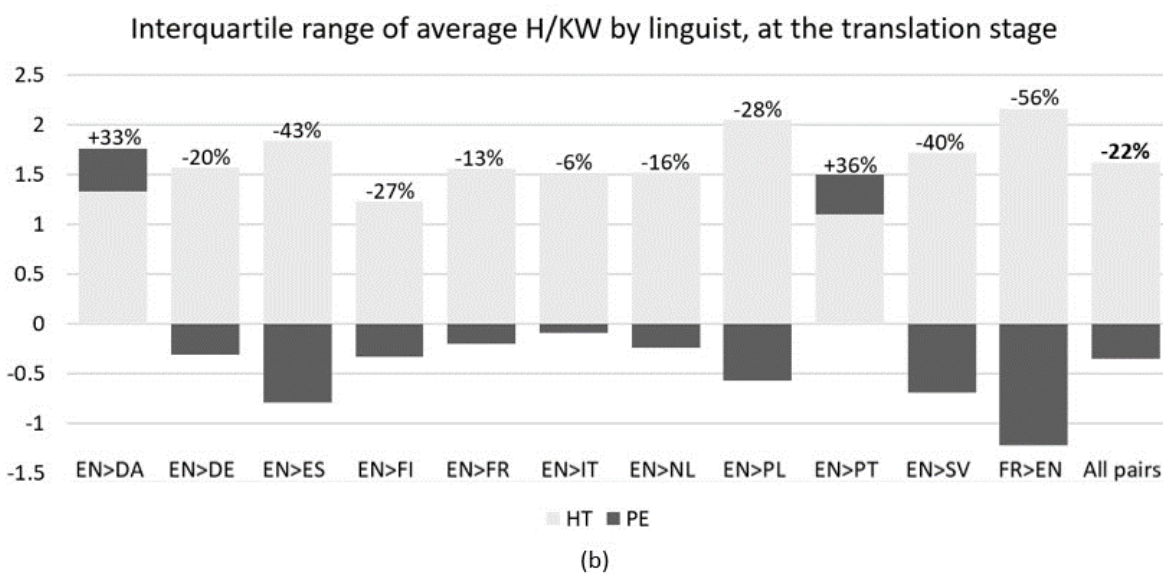

Interquartile range of average H/KW by linguist, at the translation stage

(b)

**Figure 6.** Stacked bar chart of interquartile range of average (a) WPH and (b) H/KW by linguist, at the translation stage

| | Overall dispersion of average speed by linguist, at the translation stage | | | | | |
|---|---|---|---|---|---|---|
| | WPH | | | H/KW | | |
| | HT | PE | PE-HT (%) | HT | PE | PE-HT (%) |
| IQR | 394 WPH | 616 WPH | +56% | 1.62 H/KW | 1.27 H/KW | -22% |
| SD w/ outliers | 341 WPH | 524 WPH | +54% | 4.51 H/KW | 1.78 H/KW | -58% |
| SD w/o outliers | 280 WPH | 425 WPH | +52% | 1.09 H/KW | 0.82 H/KW | -25% |

**Table 8.** Overall dispersion of average speed by linguist, at the translation stage

In line with the results of previous research (e.g. Moorkens et al. 2015; Parra Escartín and Arcedillo 2015; Toral, Wieling and Way 2018), there was high variability among the average speed obtained by different linguists, indicating that linguists' skills and practices have had a high impact on their speed. In line with the findings reported in Macken, Prou and Tezcan (2020), the dispersion among the average WPH values achieved by different linguists was much higher in PE than HT. Similarly to the values by HT/PE job, the opposite was the case when considering variability in the average H/KW values.

To consider the impact of revisers' skills and ways of working on their speed, Figure 7 presents the interquartile range of average speed by reviser, and Table 9 summarises the overall IQR and SD values.

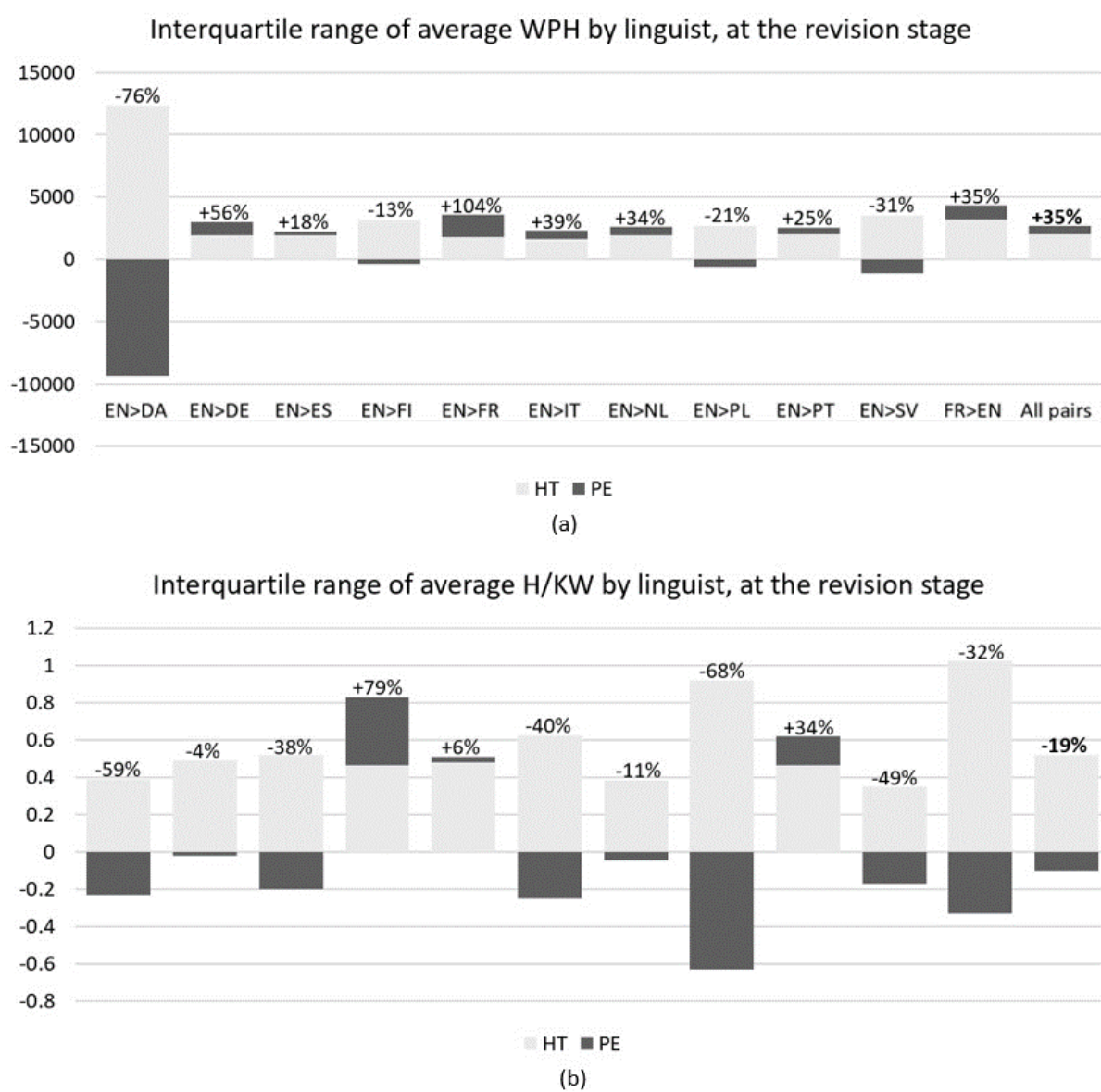

**Figure 7.** Stacked bar chart of interquartile range of average (a) WPH and (b) H/KW by linguist, at the revision stage

| | Overall dispersion of average speed by linguist, at the revision stage | | | | | |
| | WPH | | | H/KW | | |
| | HT | PE | PE-HT (%) | HT | PE | PE-HT (%) |
| IQR | 2,026 WPH | 2,730 WPH | +35% | 0.52 H/KW | 0.42 H/KW | -19% |
| SD w/ outliers | 4,183 WPH | 3,099 WPH | -26% | 0.76 H/KW | 0.48 H/KW | -37% |
| SD w/o outliers | 1,355 WPH | 1,706 WPH | +26% | 0.32 H/KW | 0.27 H/KW | -16% |

**Table 9.** Overall dispersion of average speed by linguist, at the revision stage

Once again, these values indicate that there was very high variability in the average speed obtained by different revisers, with higher dispersion of WPH values in PE, and higher variability of H/KW in HT.

All in all, high levels of variability were identified in all speed data analysed in this research. WPH variability was considerably higher in post-editing and at the revision stage, and H/KW variability was much higher in human translation and at the translation stage. I would argue that these measures of dispersion can fruitfully complement one another, as it would be highly relevant for LSPs to gain an understanding of variability not only in the number of words translated in an hour, but also in the amount of time spent to process a text of a given number of words. The high speed variability identified in this research appears to reflect the translation process, which is not as standardised and uniform as a manufacturing process; rather, it varies based on a wide range of factors, such as the features of different source texts and the translation strategies that linguists apply to deal with them, any requirements that linguists need to comply with, the quality of the MT output in the case of post-editing, etc. Investigating the impact of all these factors on speed lies beyond the scope of this article, which has focused on comparing speed variability in post-editing and human translation.

*5.3. Correlations between speed and edit distance*

To consider the extent to which post-editing speed and edit distance values are correlated, Table 10 displays the Kendall's tau-b correlations between post-editing WPH and PED, or PED-TMR, with overall values highlighted in bold.

| | Kendall's tau-b correlation between post-editing WPH and edit distance by PE job | | | | | | | |
| | PED | | | | PED-TMR | | | |
| | W/ outliers | | W/o outliers | | W/ outliers | | W/o outliers | |
| Language pair | Tau-b | p | Tau-b | p | Tau-b | p | Tau-b | p |
| EN>DA | -.353 | < .001 | -.355 | < .001 | -.068 | .014 | -.072 | .011 |
| EN>DE | -.320 | < .001 | -.297 | < .001 | -.188 | < .001 | -.162 | < .001 |
| EN>ES | -.340 | < .001 | -.331 | < .001 | -.121 | < .001 | -.120 | < .001 |
| EN>FI | -.393 | < .001 | -.377 | < .001 | -.039 | .104 | -.026 | .295 |
| EN>FR | -.379 | < .001 | -.379 | < .001 | -.253 | < .001 | -.251 | < .001 |
| EN>IT | -.440 | < .001 | -.423 | < .001 | -.297 | < .001 | -.280 | < .001 |
| EN>NL | -.321 | < .001 | -.321 | < .001 | -.207 | < .001 | -.205 | < .001 |
| EN>PL | -.356 | < .001 | -.300 | < .001 | .050 | .016 | .069 | .002 |
| EN>PT | -.321 | < .001 | -.287 | < .001 | -.363 | < .001 | -.337 | < .001 |
| EN>SV | -.268 | < .001 | -.258 | < .001 | -.031 | .268 | -.025 | .387 |
| FR>EN | -.321 | < .001 | -.254 | .002 | -.287 | < .001 | -.218 | .008 |
| All pairs | **-.367** | **< .001** | **-.355** | **< .001** | **-.191** | **< .001** | **-.183** | **< .001** |

**Table 10.** Kendall's tau-b correlation between post-editing WPH and edit distance by PE job

All values above are highly statistically significant, except for the correlations between speed and PED-TMR in English-to-Finnish and English-to-Swedish, which are not statistically significant. Overall, there was a moderate, negative correlation between speed and PED – which accounts for edits to the MT output only – whilst there was only a weak, negative correlation between speed and PED-TMR – which measures the overall technical effort. These findings indicate that there was a moderate tendency for post-editing speed to decrease as linguists made more edits to the MT output, and that the strength of this tendency dropped to weak when including corrections to TM matches and repetitions in the edit distance values. These results do not support the use of edit distance by some LSPs to determine linguists' remuneration (Albarino 2019; ELIA et al. 2022), as this metric does not correlate strongly with temporal effort.

The higher strength of the correlation between speed and PED compared to the interrelation between speed and PED-TMR may be affected by the fact that linguists working at TranslateMedia cannot edit 100% and 101% TM matches,[6] because these matches are locked. The correlation between temporal and technical effort is always 0 in these segments, and it has the potential of lowering the strength of the correlation between speed and PED-TMR for the whole text. Nevertheless, evaluating the impact of 100% and 101% TM matches on this correlation is beyond the scope of this article. The correlations identified in this research are partially in line with those reported in previous studies investigating the interrelation between speed and edit distance in NMT post-editing, which identified moderate correlations between speed and edit distance comprising changes to the MT output and TM matches (Macken, Prou and Tezcan 2020), and weak correlations when considering changes to the MT output only (Cumbreño and Aranberri 2021). However, these studies have not measured correlations between speed and edit distance both including and excluding TM matches, so we can only make limited comparisons with their findings.

---

[6] I.e. the segment in the TM is the same as the segment that needs translating; in 101% matches, one segment below and one above the selected segment are 100% matches too.

Table 11 presents the correlations between revision speed and revision distance, in HT and PE.

| | Kendall's tau-b correlation between revision WPH and revision distance by HT/PE job | | | | | | | |
|---|---|---|---|---|---|---|---|---|
| | HT | | | | PE | | | |
| | W/ outliers | | W/o outliers | | W/ outliers | | W/o outliers | |
| Language pair | Tau-b | p | Tau-b | p | Tau-b | p | Tau-b | p |
| EN>DA | -.520 | < .001 | -.488 | < .001 | -.403 | < .001 | -.374 | < .001 |
| EN>DE | -.416 | < .001 | -.401 | < .001 | -.408 | < .001 | -.385 | < .001 |
| EN>ES | -.329 | < .001 | -.304 | < .001 | -.313 | < .001 | -.289 | < .001 |
| EN>FI | -.420 | < .001 | -.406 | < .001 | -.434 | < .001 | -.410 | < .001 |
| EN>FR | -.381 | < .001 | -.335 | < .001 | -.315 | < .001 | -.279 | < .001 |
| EN>IT | -.376 | < .001 | -.346 | < .001 | -.465 | < .001 | -.452 | < .001 |
| EN>NL | -.382 | < .001 | -.340 | < .001 | -.472 | < .001 | -.425 | < .001 |
| EN>PL | -.355 | < .001 | -.300 | < .001 | -.392 | < .001 | -.347 | < .001 |
| EN>PT | -.365 | < .001 | -.320 | < .001 | -.403 | < .001 | -.357 | < .001 |
| EN>SV | -.449 | < .001 | -.424 | < .001 | -.322 | < .001 | -.299 | < .001 |
| FR>EN | -.512 | < .001 | -.492 | < .001 | -.701 | < .001 | -.640 | < .001 |
| **All pairs** | **-.397** | **< .001** | **-.359** | **< .001** | **-.395** | **< .001** | **-.365** | **< .001** |

**Table 11.** Kendall's tau-b correlation between revision WPH and revision distance by HT/PE job

With the exception of a few cases where the correlation between revision speed and revision distance was strong – i.e. French-to-English HT and PE, and English-to-Danish HT with outliers – all other language pairs presented a moderate, negative, and highly statistically significant correlation between revision speed and revision distance, both in PE and HT. This indicates that there was a moderate tendency for revision speed to decrease as revisers made more edits.

## 6. Conclusion

This article has presented the results of the first large-scale analysis of translation and revision speed in human translation and NMT post-editing, based on real-world data from an LSP. On average, PE has been 66% faster than HT, at the translation stage. However, the impact of post-editing on speed has varied dramatically among different language pairs, from an average speed increase of +130%, to a decrease of -7% – showing that PE may not always increase speed in all language combinations. Although the revision process is not supposed to differ significantly in HT or PE, on average PE revision has been 38% faster than HT revision. Overall, the most typical speed values – excluding breaks and time spent researching or reading briefing documents – have been in the ranges of 530-1,440 WPH at the translation stage of PE, and 330-820 WPH in HT; 1,990-5,540 WPH at the revision round of PE, and 1,200-3,870 in HT. These ranges provide a first reference for LSPs to compare with their speed values.

Identifying outliers through a mathematical formula has enabled me to calculate what percentage of HT/PE jobs had been time-efficient. This has been helpful especially for English-to-Polish translation: although averages including outliers suggested that PE was 18% faster than HT, identifying outliers revealed that post-editing had decreased speed in 89% of tasks. These findings highlight that average values may be misleading and are not sufficient to evaluate whether PE is faster than HT.

Besides, there was high variability in the speed obtained for different HT/PE tasks. The average speed values achieved by different linguists presented high variability too, suggesting that the skills and ways of working of individual linguists had a high impact on their speed. WPH variability was considerably higher in post-editing than in human translation, and in revision compared to the translation stage, and the opposite was the case for H/KW variability. The high levels of speed variability identified in this research appear to reflect the translation process, which is not standardised and uniform, but rather is influenced by a wide range of factors.

Additionally, in the data investigated in this research there was a moderate correlation between post-editing speed and PED (accounting for edits to the MT output), and only a weak correlation between speed and PED-TMR (which measures the overall technical effort). These findings indicate that there was a moderate tendency for post-editing speed to decrease as linguists made more edits to the MT output, and that the strength of this tendency dropped to weak when including corrections to TM matches and repetitions in the edit distance values. Moderate correlations between revision speed and revision distance were identified, both in HT and PE data. These results do not support the use of edit distance values by some LSPs to determine linguists' remuneration (ELIA et al. 2022), as this metric is not a comprehensive measure of post-editing effort.

The exploratory data analysis approach utilised in this research has provided me with a unique opportunity to investigate speed in TranslateMedia's data for 90 million translated words, yet the lack of an experimental design prevented me from controlling the variables under investigation and utilising more powerful statistical methods. Thus, it is hoped that the findings from this research may provide a foundation for future experimental analyses of speed in post-editing and human translation (Mellinger and Hanson 2016). Future studies may also seek to analyse the projects of other LSPs to confirm if revision is typically faster in PE compared to HT and, if that is the case, to investigate the reasons behind this. Besides, since the findings of this research suggest that linguists' individual skills and ways of working have a high impact on their speed, further studies could examine the skills and ways of working of different linguists, to understand which ones tend to increase or decrease speed. It would also be fruitful to determine to what extent the same linguist might spend diverse amounts of time when processing different texts, and to examine the variables behind it – such as source text features, the quality of the MT output in post-editing, etc.

In conclusion, this study has contributed to our understanding of a wide range of aspects related to speed, at the translation and revision stages of post-editing and human translation. It has also proposed an approach to evaluating speed that involves utilising an exploratory data analysis approach to account for the distribution of individual values in the analysed dataset. It is hoped that this approach will serve as a first step towards developing more comprehensive measures of speed in a professional translation context. This article also hopes to pave the way for much-needed further research into translation speed, as well as other aspects of translation productivity that are still largely underresearched.

### *Funding information*

This research was funded by the Arts and Humanities Research Council (AHRC) of UK Research and Innovation (UKRI), with grant number 2498533.

### *Acknowledgments*

I would like to express my gratitude to Prof Maeve Olohan for her invaluable guidance and feedback on my research. I would like to extend my thanks to TranslateMedia for their support and for enabling me to analyse their translation projects. I would also like to thank Dr Rebecca Tipton and Dr Henry Jones for their constructive advice.

***Author's address***

Silvia Terribile

University of Manchester

Oxford Road

Manchester, M13 9PL

UK

silvia.terribile.research@gmail.com

https://orcid.org/0000-0001-5791-926X